Title:

# On Applying Machine Learning/Object Detection Models for Analysing Digitally Captured Physical Prototypes from Engineering Design Projects

List of authors:

Jorgen F. Erichsen

Sampsa Kohtala

Martin Steinert

Torgeir Welo

Authors' Affiliation:

Norwegian University of Science and Technology (NTNU),

Department of Mechanical and Industrial Engineering

(all four authors share the same affiliation)

Corresponding Author:

Name: Jorgen F. Erichsen

Email: jorgen.erichsen@ntnu.no

Address: Richard Birkelands Veg 2B, 7491 Trondheim, Norway

Short Title:
Analysis of Prototypes by Object Detection

Number of Manuscript Pages: 13
Number of Tables: 4
Number of Figures: 3

## Abstract:

While computer vision has received increasing attention in computer science over the last decade, there are few efforts in applying this to leverage engineering design research. Existing datasets and technologies allow researchers to capture and access more observations and video files, hence analysis is becoming a limiting factor. Therefore, this paper is investigating the application of machine learning, namely object detection methods to aid in the analysis of physical porotypes.

With access to a large dataset of digitally captured physical prototypes from early-stage development projects (5950 images from 850 prototypes), the authors investigate applications that can be used for analysing this dataset. The authors retrained two pre-trained object detection models from two known frameworks, the TensorFlow Object Detection API and Darknet, using custom image sets of images of physical prototypes. As a result, a proof-of-concept of four trained models are presented; two models for detecting samples of wood-based sheet materials and two models for detecting samples containing microcontrollers.

All models are evaluated using standard metrics for object detection model performance and the applicability of using object detection models in engineering design research is discussed. Results indicate that the models can successfully classify the type of material and type of pre-made component, respectively. However, more work is needed to fully integrate object detection models in the engineering design analysis workflow. The authors also extrapolate that the use of object detection for analysing images of physical prototypes will substantially reduce the effort required for analysing large datasets in engineering design research.

# 1 Introduction and Background

With recent advances in technology, capturing various parts of design activity and design output becomes more and more feasible. However, with data sets growing in size there is a resource problem in analysing the data properly, as discussed by Törlind et al. (2009) and Erichsen, Wulvik, Steinert, & Welo (2019).

Erichsen, Sjöman, Steinert, & Welo (2019) present a method for systematically capturing physical prototypes in a digital format, and an example implementation of a capture-system for capturing prototypes is described in detail. This capture-system creates multi-view images (i.e. images from seven different viewing angles) and corresponding metadata (including information about when, where and by whom the prototype was captured). The method aims at enabling researchers by providing more observations (i.e. captured prototypes) of early-stage design activities that can be analysed. This proposed method of capturing prototypes is somewhat similar to efforts capturing later stage design activities, e.g. Computer-Aided Design (CAD) (Ishino & Jin, 2002; Levy, Rafaeli, & Ariel, 2016), yet focuses on capturing (physical) output from early-stage physical prototyping.

With the increasing focus and progress made towards image processing and object detection and recognition within data science and machine learning, using (and repurposing) existing methods and models for object detection has become increasingly more available. Specifically, Convolutional Neural Networks (CNNs) can be trained using large quantities of images for recognizing patterns (Zeiler & Fergus, 2014). With a growing community of researchers focusing on better performing models for object detection and recognition, it is possible to use existing models and to perform a retraining of the final layers—essentially repurposing the model to handle new image data (i.e. 'classes' of objects). Object detection models are often benchmarked using commonly accepted tests, e.g. the ImageNet benchmark (Krizhevsky, Sutskever, & Hinton, 2012), and researchers wanting to retrain these models must often choose between fast (real-time) processing speed and prediction accuracy (Huang et al., 2016).

By capturing 850 prototypes (as of March 20$^{th}$ 2019) from ongoing early-stage PD projects using the capture system presented by (Erichsen, Sjöman, et al., 2019), the authors and colleagues have demonstrated that capturing physical prototypes from ongoing projects is both possible and feasible, and that the captured data can be used for analysis of output from PD projects. However, the resources required for analysing the captured data are substantial. This is a problem similar to the analysis problem faced by researchers doing manual video coding of design observations (Törlind et al., 2009). Consequently, it is of great interest to explore different tools that can be applied to solve this analysis problem. Whereas Erichsen, Sjöman, et al. (2019) present manual categorisations of various properties of captured prototypes, e.g. materials and tools used to make the prototypes, this article investigates automatically classifying digitally captured physical prototypes using object classification.

The aim of this article is to explore and discuss what is possible, feasible and applicable when using object detection to analyse physical prototypes that are digitally captured from PD projects. The article addresses the following two research questions:

RQ1. Can materials be automatically classified from images of physical prototypes, and is it feasible to use this for analysis of captured prototypes from PD projects?

RQ2. Can pre-made components be automatically classified from images of prototypes, and is it feasible to use this for analysis of captured prototypes from PD projects?

As the aim of this article is to explore categorising digitally captured physical prototypes, the authors have decided to use object detection for doing this classification. Object detection has been chosen over image classification because of the possibility of determining size and position of objects in the images. Additionally, object detection can classify multiple objects per image.

This article explores analysing prototypes by retraining two pre-trained object detection models from two known frameworks, using custom image datasets of images of physical prototypes. The article focuses on retraining pre-trained models with custom datasets as a proof-of-concept, as using known frameworks can make this approach available to more researchers. This is prioritized over achieving state-of-the-art accuracy during inference for the different models.

To simplify matters somewhat—since this is a proof-of-concept—the authors have chosen to classify one type of materials and one type of pre-made components, both which are often present in prototypes from the dataset presented by Erichsen, Sjöman, et al. (2019). Arguably, image recognition would be suitable for this specific application—i.e. classifying a single object from an image. Yet, the intended use-case is to classify more than one material and more than one type of pre-made component per image—which would be a more suitable application for object detection.

The type of materials chosen is wood-based sheet materials, e.g. Medium Density Fibreboard (MDF) and plywood. Many of the prototypes presented by Erichsen, Sjöman, et al. (2019) include such materials, as these are inexpensive materials that are relatively easy to work with. Wood-based sheet materials like MDF are quite distinct in texture and colour, which should make such materials easier to differentiate. However, since these materials are independent of shape, this may complicate the detection process.

The pre-made component of choice has been microcontrollers, e.g. the various types of Arduinos ('Arduino', 2019; Barrett, 2013) available. In contrast to wood-based sheet materials, the microcontrollers are relatively homogenous in shape, size and colour (albeit with some variation). However, they are much smaller objects in general, which might make them much harder to detect in lower-resolution images. Moreover, from an engineering design research standpoint, identifying the use of microcontrollers in early-stage lower resolution prototypes is especially interesting because it might indicate a jump in functionality and interactivity provided by the prototype.

## 2 Method

As this article explores a proof-of-concept for classifying images of physical prototypes, object detection has been chosen as the preferred classification method. For training the models, the authors have chosen to use two popular object detection frameworks; the TensorFlow Object Detection API (Huang et al., 2016), and Darknet (Redmon, 2013).

### 2.1 Choice of Models

Since the two frameworks, TensorFlow Object Detection API (Huang et al., 2016) and Darknet (Redmon, 2013), have different implementations of the various machine learning models, a separate pre-trained model has been chosen for each framework. Effectively, this means that the authors have retrained four pre-trained models for classifying objects in images of physical prototypes:

Model A   Classifying wood-based sheet materials through retraining the Faster-RCNN model by using the TensorFlow Object Detection API framework

Model B   Classifying wood-based sheet materials through retraining the darknet53 model by using the Darknet framework.

Model C    Classifying microcontrollers through retraining the Faster-RCNN model by using the TensorFlow Object Detection API framework.

Model D    Classifying microcontrollers through retraining the darknet53 model by using the Darknet framework.

The pre-trained models used in this article are Faster-RCNN (Ren, He, Girshick, & Sun, 2015) and darknet53 (from its 53 convolutional layers) and is the backbone of YOLOv3 (Redmon & Farhadi, 2018). Both of these models have been chosen because they sport a good balance between training speed and prediction accuracy (Huang et al., 2016; Redmon & Farhadi, 2018). The pre-trained Faster-RCNN (Ren et al., 2015) model used in this article has been trained on the Inception v2 dataset (Ioffe & Szegedy, 2015), and the pre-trained darknet53 model has been trained on the ImageNet (Russakovsky et al., 2015) dataset.

By default, the TensorFlow Object Detection API uses a confidence threshold of 0.5, meaning that a proposed bounding box must have a confidence of more than 0.5 to count as a predicted object. Darknet uses a default confidence threshold of 0.25, yet this was modified to 0.5 to get a fair comparison between the two models (and for reducing the potential number of false positives of the YOLOv3 model during inference).

### 2.2   Custom Training, Validation and Test Datasets

This article uses custom datasets for retraining the four models. A large portion of the data used in this article has been captured using the system described by Erichsen, Sjöman, et al. (2019). This capture system is shown to the left in Figure 1 and features seven Full HD (i.e. 1920 by 1080 pixels) cameras that capture different viewing angles of the prototypes. The corresponding output from this system, i.e. a captured physical prototype, is shown to the right in Figure 1.

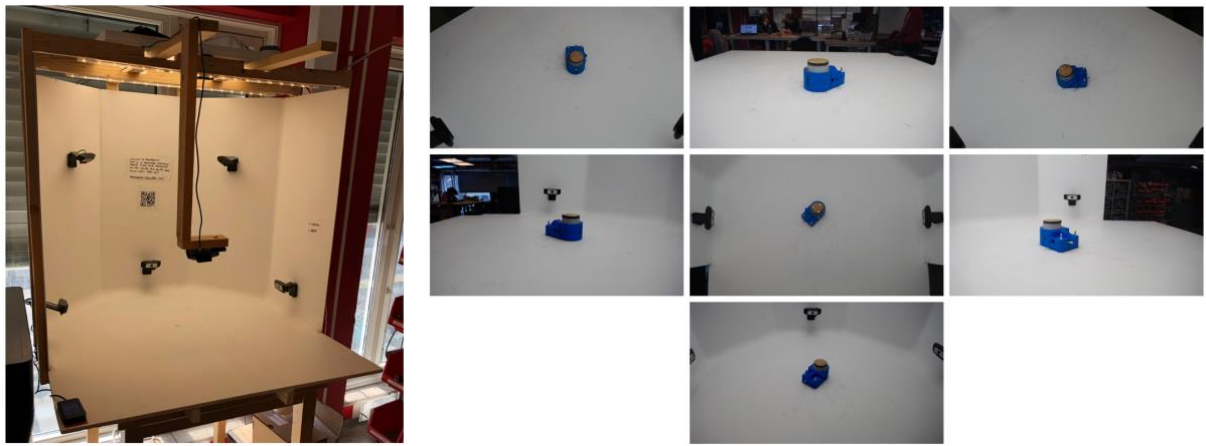

Figure 1 The physical capture system used for capturing the prototypes (left) together with an example of a multi-view image of one prototype (right), where seven viewing angles are captured by Full HD cameras.

By capturing a single prototype from seven different viewing angles, details from different sides of the prototype are easier to capture. Similarly, some parts of the prototype may be obscured in from one viewing angle, while visible from another. This also means that capturing one prototype, there are seven possible angles which can be used for (either) training or inference.

Most of the images of prototypes used in this article are of prototypes captured during early-stage PD projects. Additionally, to add some diversity to the datasets, images from Google Images ('Google Images', 2019) have been added, as well as some material samples captured by the system described by Erichsen, Sjöman, et al. (2019).

In an attempt to further increase the stability of the models, negative examples are included when training, evaluating and testing all models. Most of the negative samples are images from an "empty" capture system—as shown to the left in Figure 1—but also images of prototypes that do not contain wood-based sheet materials or microcontrollers. Since the images from the "empty" capture system images are taken from cameras with fixed positions, they appear very similar, with only slight variations to light (due to differences in time-of-capture) and exposure (due to differences in lighting, as the exposure of the used cameras is set automatically). The same number of "empty" images were used for training all four models.

The authors have gathered a total of 1624 images for classifying wood-based sheet materials and 1273 images for classifying microcontrollers. The total number of objects labelled (as there could be more than one labelled object per image) were 1426 for wood-based sheet materials and 928 for microcontrollers. The two sets of images have been split into training, validation and test sets, as shown in Table 1.

Table 1 Distribution of images in the custom training, validation and test sets.

|  | Model A and B (Wood-Based Sheets) | | | Model C and D (Microcontrollers) | | |
| --- | --- | --- | --- | --- | --- | --- |
| Dataset | Training | Validation | Test | Training | Validation | Test |
| Total number of images | 1243 (76.5%) | 134 (8.3%) | 247 (15.2%) | 863 (67.8%) | 217 (17.0%) | 195 (15.3%) |
| Total labelled objects (regardless of files) | 1222 (85.7%) | 88 (6.2%) | 116 (8.1%) | 653 (70.4%) | 169 (18.2%) | 106 (11.4%) |

Of all the images for classifying wood-based sheet materials, 51.8% of the images included positive samples (i.e. contains wood-based sheet materials), and of all the images for classifying microcontrollers, 55.7% of the images included positive samples (i.e. contains microcontrollers). The portions of images with positive and negative samples and how they are distributed throughout the training, validation and test sets are shown in Table 2.

Table 2 Portion of images with positive and negative samples for each dataset.

|  | Model A and B (Wood-Based Sheets) | | | Model C and D (Microcontrollers) | | |
| --- | --- | --- | --- | --- | --- | --- |
| Dataset | Training | Validation | Test | Training | Validation | Test |
| Images with positive samples | 682 (54.87%) | 62 (46.27%) | 98 (39.68%) | 518 (60.02%) | 101 (46.54%) | 90 (46.15%) |
| Images with negative samples | 561 (45.13%) | 72 (53.73%) | 149 (60.32%) | 345 (39.98%) | 116 (53.46%) | 105 (53.85%) |

## 2.3 Hardware Used for Training

To demonstrate that there are multiple approaches to training the models presented in this article, the authors have trained two of the models locally using a personal computer, and two of the models online using Google Cloud's ML-engine platform. Since the two frameworks used have different implementations, the models trained with TensorFlow Object Detection API (i.e. Model A and Model C) were trained online and the models trained with Darknet (i.e. Model B and Model D) were trained locally. Notably, both frameworks have cloud-based and local implementations, and elapsed time required for training the models will vary with training parameters and hardware used for both cloud-based and local implementations.

During cloud training, the hardware used was Google Cloud's standard machine type with 8 virtual Central Processing Units (vCPUs), 30 GB of memory and a Tesla P100 Graphics Processing Unit (GPU). Local training was performed on a laptop with a 6-core processor (Intel

i9-8950HK), 32 GB of memory and an external graphics enclosure with an NVIDIA GeForce RTX 2080 Ti GPU.

## 2.4 Training Parameters

Before training the four models, training parameters were decided for each of the frameworks. For the two models trained using TensorFlow Object Detection API, Model A and Model C, a fixed learning rate of 0.000001 was chosen, along with a batch size of 1. Both Model A and Model C were set to train for a maximum of 400 000 steps on the corresponding training datasets, with evaluation on the corresponding validation sets for every epoch. During training, a random number of images was flipped horizontally—which was the only pre-processing or augmenting applied to the images. All images were kept at their original aspect ratio and resized to match a resolution between minimum 600 pixels and maximum 1024 pixels vertically.

For training Model B and Model D with the Darknet framework, the learning rate was set to 0.001 with a batch size of 64. The batch was further subdivided into 16 blocks for parallel computation on the GPU. Model B and Model D were set to train indefinitely, requiring the authors to manually stop the training. Evaluation on the corresponding validation sets were done for every 4th epoch for Model B and Model D. Each image was resized to fit within the network resolution set to maximum 416 pixels in both width and height, while still keeping the original aspect ratio.

It is worth noting that since the size of an epoch (i.e. the iteration over the entire set) is dependent on the number of images and the batch size, the number of (training) steps per epoch will be different for all four models. Therefore, the number of (training) steps per epoch is 1243 for Model A, 20 for Model B, 863 for Model C and 14 for Model D.

## 2.5 Metrics for Evaluating Performance

Mean Average Precision (mAP) is a known metric for evaluating object detection performance (Everingham et al., 2015; Everingham, Van Gool, Williams, Winn, & Zisserman, 2010; Russakovsky et al., 2015), especially when comparing the performance of models on the same datasets. The mAP value of a set of predictions is the mean over classes of the interpolated Average Precision (AP) (Salton & McGill, 1986) for each class (Henderson & Ferrari, 2016). The mAP is dependent on the per-class precision-recall curve, and therefore also the area of which the predicting bounding box overlaps with the ground truth labelled object—the Intersection-over-Union (IoU) (Everingham et al., 2015). An IoU of 0.75 means that there is a 75% overlap between the predicted label and the ground truth label. Similarly, an IoU of 0.5 indicates a 50% overlap. Hence, mAP values for various IoU thresholds are good metrics for not only assessing if a model performs well in recognising a type of object, but also for assessing the model's ability to pinpoint the object position in an image. Therefore, this article uses mAP at 0.5 and 0.75 IoU thresholds as the most important metrics for evaluating the performance of the four models. It's worth noting that a model's mAP will change depending on the test set, so the mAP values of all four models will not be directly comparable—mAP of Model A and Model B can be compared, as these share the same test set. The same applies to Model C and Model D.

Since this proof-of-concept also attempts to identify the existence of a single object in the image, and not only assessing the size or number of times the object is included, it is also interesting to evaluate how the trained models perform at this task. Therefore, assessing the per-image precision/recall and accuracy (as opposed to the per-class) is also of interest, and is included as a supplementary metric for evaluating the performance of the models.

## 3 Results

Training of all four models was performed in March 2019. Model A and Model C were trained using TensorFlow Object Detection API on Google Cloud's ML-engine platform, while Model B and Model D were trained locally using Darknet. Model A and Model C were trained for approximately 16 hours each before reaching the set limit of 400 000 steps—which is equivalent to approximately 321 and 463 epochs, respectively. Model B was trained for approximately 9 hours before being stopped after 14 700 epochs (which equates to approximately 294 000 steps for Model B), when the average loss had reached 0.0449. Model D trained for approximately 6,5 hours before being stopped after 11000 epochs (which equates to approximately 154 000 steps for Model D), when the average loss had reached 0.0791.

Table 3 Resulting mAP values at 0.5 and 0.75 IoU thresholds for the four models trained.

|  | Model A (Wood-Based Sheet Materials) | | Model B (Wood-Based Sheet Materials) | | Model C (Microcontrollers) | | Model D (Microcontrollers) | |
| --- | --- | --- | --- | --- | --- | --- | --- | --- |
| Dataset | Validation | Test | Validation | Test | Validation | Test | Validation | Test |
| mAP @ 0.5 IoU | 0.684 | 0.521 | 0.905 | 0.766 | 0.741 | 0.649 | 0.804 | 0.681 |
| mAP @ 0.75 IoU | 0.416 | 0.280 | 0.704 | 0.399 | 0.435 | 0.351 | 0.603 | 0.369 |

All four models were evaluated on their corresponding validation sets during training. After training, mAP numbers for IoU thresholds of 0.5 and 0.75 were collected for the corresponding validation and test sets, and are shown in Table 3. The two models trained for classifying wood-based sheet materials, Model A and Model B, achieved mAP values at 0.5 IoU of 0.521 (for Model A) and 0.766 (for Model B) and mAP values at 0.75 IoU of 0.280 (for Model A) and 0.399 (for Model B) when evaluated on the test set. The models trained for classifying microcontrollers, Model C and Model D, achieved mAP values at 0.5 IoU of 0.649 (for Model C) and 0.681 (for Model D) and mAP values at 0.75 IoU of 0.351 (for Model C) and 0.369 (for Model D) when evaluated on the test set. An example from a successful detection of a microcontroller by Model C is shown in Figure 2.

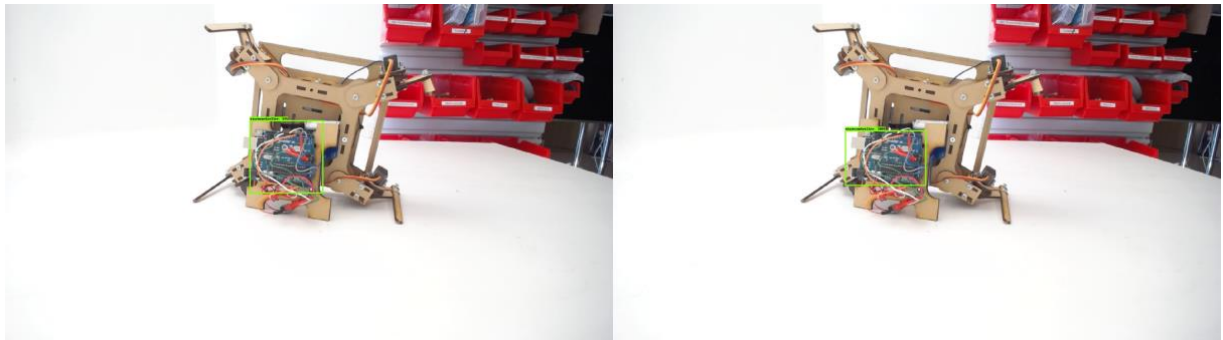

Figure 2 Image with successful detection of a microcontroller from Model C (left) and ground truth label (right) from the test set for classifying microcontrollers.

Additionally, the authors performed a manual, per-image evaluation of the four models using the corresponding test sets, categorising the images as either true positives, false positives, true negatives or false negatives. From these evaluations, the per-image precision, recall and accuracy of the four models were calculated, and are shown in Table 4.

Table 4 Per-image precision, recall and accuracy for all four trained models evaluated on their corresponding test sets, with evaluation done per image rather than per object.

|  | True Positives | False Positives | True Negatives | False Negatives | Precision | Recall | Accuracy |
|---|---|---|---|---|---|---|---|
| Model A (Wood-Based Sheet Materials) | 72 | 4 | 168 | 3 | 0.947 | 0.960 | 0.972 |
| Model B (Wood-Based Sheet Materials) | 90 | 8 | 141 | 8 | 0.918 | 0.918 | 0.935 |
| Model C (Microcontrollers) | 60 | 5 | 102 | 28 | 0.923 | 0.682 | 0.831 |
| Model D (Microcontrollers) | 72 | 0 | 101 | 22 | 1.000 | 0.766 | 0.887 |

# 4 Interpretation of Results

From investigating Table 3, it is apparent that all four models perform worse on their respective test sets compared to their respective validation sets. This is to be expected, as the validation set is used to update the models' hyperparameters during training, whereas the test set is left unused until after training has finished (i.e. unbiased). Also as expected, all models have considerably lower mAP values at 0.75 IoU than at 0.5 IoU.

Comparing Model A and Model B shows that Model B performs better than Model A at both 0.5 and 0.75 IoU. Similarly, Model D outperforms Model C by a slight margin at both 0.5 and 0.75 IoU. This might be due to the different training parameters set for the models, e.g. batch size, resolution and learning rate, as these were set differently for Model A and Model C than for Model B and Model D.

On manual inspection of the evaluations (on both validation and test sets), Model A sometimes produces several positive predictions per object, as shown in Figure 3. This might explain the relatively low mAP at 0.75 IoU for Model A, even though the model correctly recognises the existence of the material in the image.

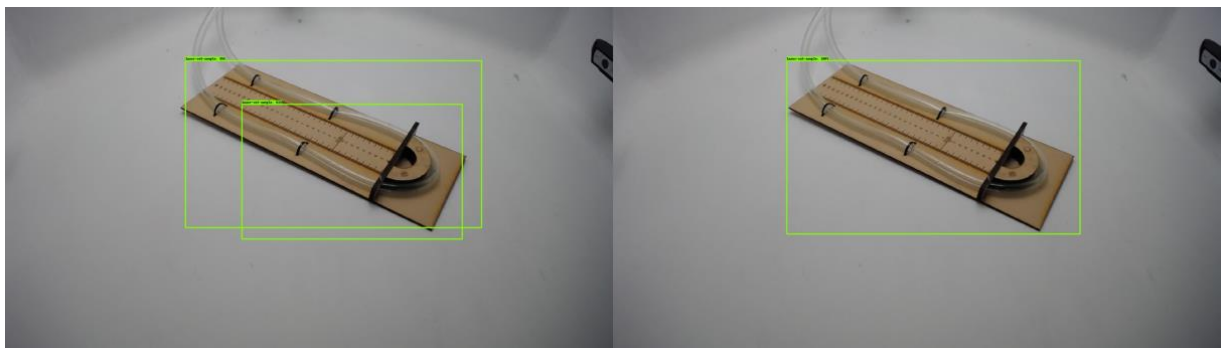

Figure 3 Image with predicted labels from Model A (left) and ground truth label (right) from the test set for classifying wood-based sheet materials.

The main goal of the per-image evaluation is to assess the four models' abilities in identifying the existence of the object in the image, not assessing the size or number of times the object is included. Upon inspecting the per-image evaluation presented in Table 4, it is apparent that all four models have high precision values—i.e. there are few false positives. This is perhaps more surprising for Model A and Model B, as the wood-based sheet material samples used in the dataset have many varying shapes and sizes—as opposed to the microcontroller samples, which are quite similar in shape and size. The relatively low recall for Model C and Model D indicate that these models are more conservative, which is apparent because they have more false negatives. This might be explained from most of the microcontroller samples being relatively small, i.e. covering few pixels per image, and therefore sometimes hard to identify.

All four models have a relatively high number of true negatives, which is a result of the models successfully not labelling the negative samples included in the datasets. It is worth noting that while the negative samples do not affect the per-image precision and recall, they do affect per-image accuracy and also the previously discussed mAP values.

## 5 Assessing Present and Future Applicability for Engineering Design Research

The aim of this article has been to explore and discuss what is possible, feasible and applicable when using object detection to analyse digitally captured physical prototypes. The results from training the four models presented in this article show that Model A and Model B are able to successfully identify wood-based sheet materials and that Model C and Model D are able to successfully identify microcontrollers in images of physical prototypes. However, this proof-of-concept shows that the accuracy (and subsequently, the mAP) of the models could be further improved.

As this proof-of-concept has only included one class per model, a natural next step would be to train models with more classes. Because the included models are only trained to detect one class each, the research questions of this article, RQ1 ("Can materials be automatically classified from images of physical prototypes, and is it feasible to use this for analysis of captured prototypes from PD projects?") and RQ2 ("Can pre-made components be automatically classified from images of prototypes, and is it feasible to use this for analysis of captured prototypes from PD projects?"), are only partially answered. Using the four models presented in this article for analysing images of physical prototypes is both possible and feasible yet the authors note that adding more classes would substantially increase the applicability and value of the models.

For engineering design researchers that wish to apply pre-trained object detection models in their workflow, there are tutorials and extensive walkthroughs on several popular frameworks available online. However, if the aim is to detect other classes than the ones built into the pre-trained models, it is necessary to retrain the pre-trained models with custom datasets. Gathering and labelling these datasets can be quite laborious, and the general recommendation is that more data leads to better model accuracy and stability during inference. Though the models in this article have been trained to identify one class per model, the models can be retrained to detect a number of different classes each. However, when training multiple classes per model, more training data is needed.

The authors do believe that with more time and effort spent in training the models using larger datasets and more classes, this way of analysing images of physical prototypes will become applicable to a large number of engineering design researchers. While gathering data and training the models can be a laborious process—especially for engineering design researchers with limited experience with programming, artificial intelligence and object detection— performing inference on images with the trained models is relatively easy in comparison. Performing inference on a single image takes a few milliseconds, making it possible to analyse a large number of images (or even video) in a short amount of time. Therefore, the authors deem it feasible to employ object detection for analysing images of physical prototypes in engineering design research, and that it is an important step towards overcoming the problem of resources required for analysis, as discussed by Törlind et al. (2009) and Erichsen, Wulvik, et al. (2019). The models trained and tested in this article are intended for classifying objects within images of physical prototypes and have not been tested outside of this context.

# 6  Conclusion

While object detection and computer vision has received increasing attention in computer science over the last decade, there are few efforts in applying this to leverage engineering design research. This article has presented a proof-of-concept for classifying one type of materials and one type of pre-made components in images of physical prototypes. This proof-of-concept includes retraining pre-trained models for object detection that have been retrained with custom datasets.

Findings from this article indicate that the models are able to successfully classify the type of material and type of pre-made component, respectively. However, more work is needed to fully integrate object detection in the engineering design analysis workflow.

Finally, the authors extrapolate that the use of object detection for analysing images of physical prototypes might contribute to substantially reducing the effort required for analysing large datasets in engineering design research.

# 7  Acknowledgment

This research is supported by the Research Council of Norway through its user-driven research (BIA) funding scheme, project number 236739.

# 8 References

Arduino. (2019). Retrieved 24 April 2019, from https://www.arduino.cc/

Barrett, S. F. (2013). Arduino microcontroller processing for everyone! Synthesis Lectures on Digital Circuits and Systems, 8(4), 1–513.

Erichsen, J. F., Sjöman, H., Steinert, M., & Welo, T. (2019). Digitally Capturing Physical Prototypes During Early-Stage Product Development Projects for Analysis. Artificial Intelligence for Engineering Design, Analysis and Manufacturing: AI EDAM; Cambridge. Submitted, in review.

Erichsen, J. F., Wulvik, A., Steinert, M., & Welo, T. (2019). Efforts on Capturing Prototyping and Design Activity in Engineering Design Research. Procedia CIRP.

Everingham, M., Eslami, S. A., Van Gool, L., Williams, C. K., Winn, J., & Zisserman, A. (2015). The pascal visual object classes challenge: A retrospective. International Journal of Computer Vision, 111(1), 98–136.

Everingham, M., Van Gool, L., Williams, C. K., Winn, J., & Zisserman, A. (2010). The pascal visual object classes (voc) challenge. International Journal of Computer Vision, 88(2), 303–338.

Google Images. (2019). Retrieved 24 April 2019, from https://www.google.com/imghp

Henderson, P., & Ferrari, V. (2016). End-to-end training of object class detectors for mean average precision. Asian Conference on Computer Vision, 198–213. Springer.

Huang, J., Rathod, V., Sun, C., Zhu, M., Korattikara, A., Fathi, A., … Murphy, K. (2016). Speed/accuracy trade-offs for modern convolutional object detectors. ArXiv:1611.10012 [Cs]. Retrieved from http://arxiv.org/abs/1611.10012

Ioffe, S., & Szegedy, C. (2015). Batch normalization: Accelerating deep network training by reducing internal covariate shift. ArXiv Preprint ArXiv:1502.03167.

Ishino, Y., & Jin, Y. (2002). Acquiring engineering knowledge from design processes. AI EDAM, 16(2), 73–91. https://doi.org/10.1017/S0890060402020073

Krizhevsky, A., Sutskever, I., & Hinton, G. E. (2012). Imagenet classification with deep convolutional neural networks. Advances in Neural Information Processing Systems, 1097–1105.

Levy, E. C., Rafaeli, S., & Ariel, Y. (2016). The effect of online interruptions on the quality of cognitive performance. Telematics and Informatics, 33(4), 1014–1021. https://doi.org/10.1016/j.tele.2016.03.003

Redmon, J. (2013). Darknet: Open Source Neural Networks in C. Retrieved from http://pjreddie.com/darknet/

Redmon, J., & Farhadi, A. (2018). YOLOv3: An Incremental Improvement. ArXiv:1804.02767 [Cs]. Retrieved from http://arxiv.org/abs/1804.02767

Ren, S., He, K., Girshick, R., & Sun, J. (2015). Faster R-CNN: Towards Real-Time Object Detection with Region Proposal Networks. In C. Cortes, N. D. Lawrence, D. D. Lee, M. Sugiyama, & R. Garnett (Eds.), Advances in Neural Information Processing Systems 28 (pp. 91–99). Retrieved from http://papers.nips.cc/paper/5638-faster-r-cnn-towards-real-time-object-detection-with-region-proposal-networks.pdf

Russakovsky, O., Deng, J., Su, H., Krause, J., Satheesh, S., Ma, S., … Bernstein, M. (2015). Imagenet large scale visual recognition challenge. International Journal of Computer Vision, 115(3), 211–252.

Salton, G., & McGill, M. J. (1986). Introduction to modern information retrieval.

Törlind, P., Sonalkar, N., Bergström, M., Blanco, E., Hicks, B., & McAlpine, H. (2009). Lessons learned and future challenges for design observatory research. DS 58-2: Proceedings of ICED 09, the 17th International Conference on Engineering Design, Vol. 2, Design Theory and Research Methodology, Palo Alto, CA, USA, 24.-27.08. 2009.

Zeiler, M. D., & Fergus, R. (2014). Visualizing and Understanding Convolutional Networks. In D. Fleet, T. Pajdla, B. Schiele, & T. Tuytelaars (Eds.), Computer Vision – ECCV 2014 (pp. 818–833). Retrieved from https://arxiv.org/pdf/1311.2901.pdf